\documentclass{article}
\usepackage{spconf,amsmath,graphicx,hyperref}
\hypersetup{colorlinks=false, pdfborder={0 0 0}}

\title{VTONGuard: Automatic Detection and Authentication of AI-Generated Virtual Try-On Content}
\name{\shortstack{Shengyi Wu$^{1}$, Yan Hong$^{2}$, Shengyao Chen$^{1}$, Zheng Wang$^{1}$, \\ Xianbing Sun$^{1}$, Jiahui Zhan$^{1}$, Jun Lan$^{2}$\sthanks{Corresponding authors.}, Jianfu Zhang$^{1}$\footnotemark[1]}}
\address{$^{1}$ Shanghai Jiao Tong University, 
$^{2}$ Ant Group
\\ \small wsykk2@sjtu.edu.cn, c.sis@sjtu.edu.cn, lanjun\_yelan@163.com}

\begin{document}
\ninept
\maketitle
\begin{abstract}

With the rapid advancement of generative AI, virtual try-on (VTON) systems are becoming increasingly common in e-commerce and digital entertainment. However, the growing realism of AI-generated try-on content raises pressing concerns about authenticity and responsible use. To address this, we present VTONGuard, a large-scale benchmark dataset containing over 775,000 real and synthetic try-on images. The dataset covers diverse real-world conditions, including variations in pose, background, and garment styles, and provides both authentic and manipulated examples. Based on this benchmark, we conduct a systematic evaluation of multiple detection paradigms under unified training and testing protocols. Our results reveal each method’s strengths and weaknesses and highlight the persistent challenge of cross-paradigm generalization. To further advance detection, we design a multi-task framework that integrates auxiliary segmentation to enhance boundary-aware feature learning, achieving the best overall performance on VTONGuard. We expect this benchmark to enable fair comparisons, facilitate the development of more robust detection models, and promote the safe and responsible deployment of VTON technologies in practice.
\end{abstract}

\section{Introduction}


Virtual try-on (VTON) systems have become transformative in fashion, e-commerce, and digital entertainment, enabling users to preview garments through photorealistic synthesis from garment-person pairs. As demand for immersive shopping experiences grows, VTON technologies are increasingly positioned as a bridge between online and in-store retail.
The core challenge in VTON lies in achieving precise garment-body alignment while preserving fine-grained texture fidelity under diverse poses and clothing styles~\cite{lee2022high,morelli2022dress,morelli2023ladi}. Over the past few years, approaches have evolved significantly from early flow-based warping methods~\cite{wang2018toward,ge2021parser} that explicitly estimate pixel correspondences, to adversarial frameworks~\cite{choi2021viton,xie2023gp} leveraging discriminators for sharper synthesis, and more recently to diffusion-based models~\cite{yang2023paint,gou2023taming}  capable of generating highly detailed and coherent results. These innovations have not only improved alignment accuracy and texture realism but also enhanced robustness to challenging factors such as large pose variations, complex garment patterns, and diverse background conditions. Current methods~\cite{Choi2024ImprovingDM,Sun2025DSVTONHV} already achieve photorealistic quality that closely approaches real images and continue to evolve rapidly, fueled by breakthroughs in model architectures and the growing availability of diverse, large-scale datasets. This rapid evolution is steadily closing the gap between synthetic and real try-on images, paving the way for new opportunities and new challenges in their practical deployment.

Despite these advances, VTON also raises pressing concerns about authenticity and security. As generative models increasingly produce try-on images that are nearly indistinguishable from real photographs, the line between real and synthetic content is blurring~\cite{Nguyen2024SwiftTryFA,Li2025MagicTryOnHD,Zhang2024BooWVTONBI}. This introduces risks of misinformation, deceptive advertising, and misleading consumer experiences, further complicated by the diversity of generative models and their distinct artifacts.
Addressing these challenges requires reliable detectors to distinguish real from synthetic try-on images~\cite{wang2020cnn,liu2021Swin,guo2023hierarchical}. VTON presents unique difficulties: garments must align naturally with human bodies and preserve fine-grained textures, making generated results highly convincing. Existing detection datasets and methods~\cite{Wang2022DiffusionDBAL,Zhu2023GenImageAM,wang2025opensdi}, such as camera/JPEG~\cite{Kwon2021LearningJC} and diffusion-trace~\cite{Lei2024DiffuseTraceAT} approaches, though effective for general AI-generated content, are not tailored to VTON, where the subtle visual gap between real and synthetic images and the complexity of garment-body interactions pose distinct challenges.

\begin{figure}[t]
  \centering
  \includegraphics[width=\linewidth]{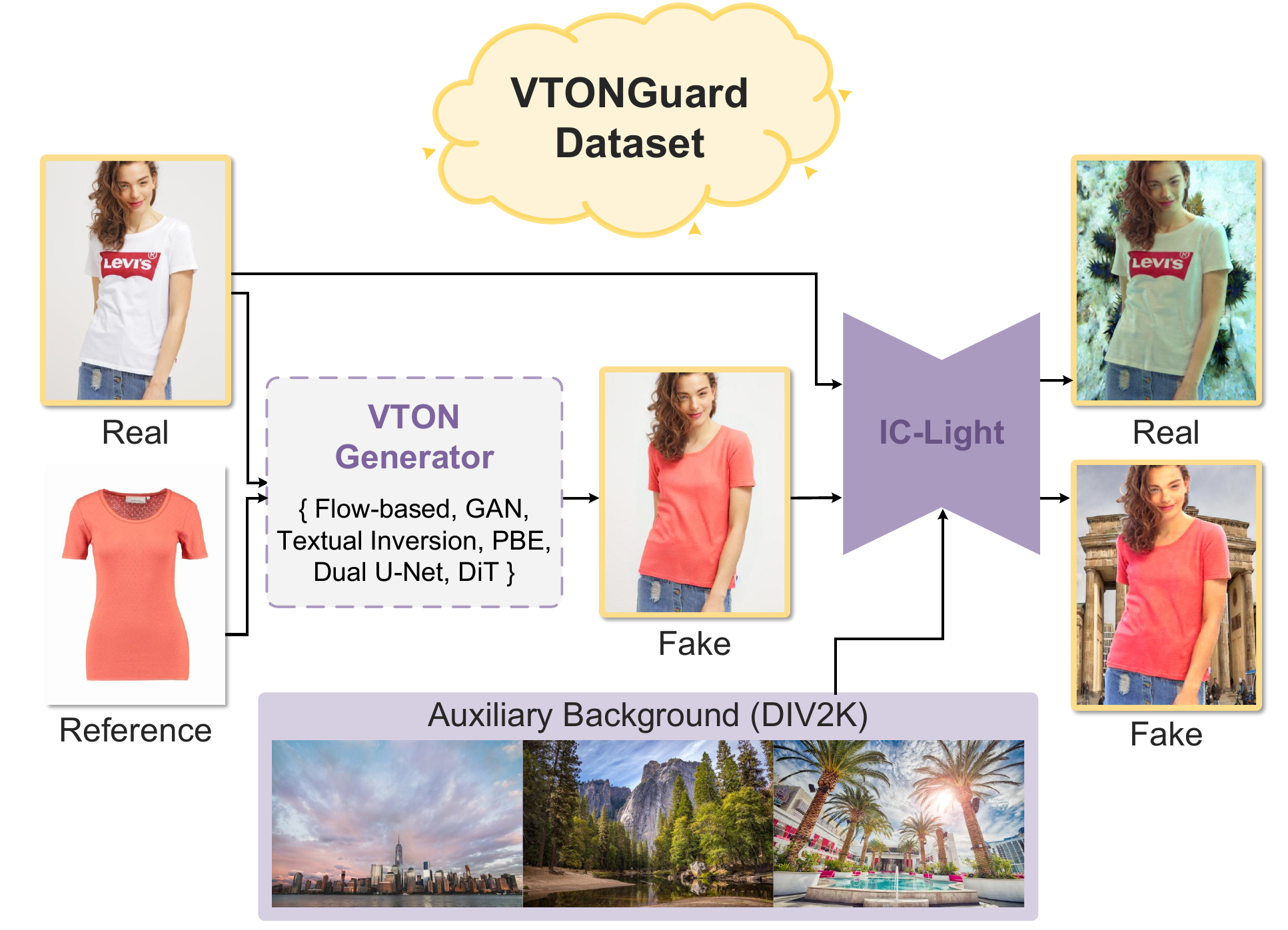}
  \caption{Overall construction pipeline of the proposed \textit{VTONGuard} benchmark, showing real and synthetic data collection, IC-Light harmonization, generative paradigm partitioning, and background augmentation with DIV2K for realistic VTON detection. }
  \label{fig:VTONGuard_pipeline}
\end{figure}

To address these challenges, we present VTONGuard, a large-scale benchmark for detecting AI-generated virtual try-on content under real-world conditions. The dataset contains 775,105 images (real/synthetic, 377,144/397,961) spanning diverse poses, garments, and backgrounds, capturing the complexity of in-the-wild garment-body alignment and fine-grained textures often missed by prior controlled datasets.
Based on this benchmark, we conduct systematic evaluations of convolutional, transformer-based, and frequency-domain detectors. We further introduce MiT-B2-MT, a multi-task hierarchical transformer with an auxiliary segmentation branch that explicitly models garment-body boundaries and stitching cues. This design achieves state-of-the-art performance across diverse generative paradigms, setting a new standard for VTON authenticity detection.

\section{VTONGuard}

The construction pipeline of the proposed \textit{VTONGuard} benchmark is illustrated in Fig.~\ref{fig:VTONGuard_pipeline}. Starting from multiple real person--garment pairs and reference backgrounds, we first generate synthetic try-on images across diverse paradigms. We further augment the dataset with auxiliary background data to improve realism and diversity, and then harmonize visual conditions using IC-Light for better background/illumination consistency. Finally, the resulting real and synthetic images are partitioned into subsets.


\subsection{VTON Models}
To comprehensively evaluate model generalization across diverse VTON paradigms, we divide our benchmark into six subsets that reflect key methodological trends: flow-based~\cite{wang2018toward,ge2021parser}, GAN~\cite{lee2022high,choi2021viton,xie2023gp}, textual inversion~\cite{morelli2023ladi}, paint-by-example (PBE)~\cite{yang2023paint,gou2023taming}, dual U-Net~\cite{zhou2024learningFF,hu2024animate}, and DiT~\cite{peebles2023scalable,Jiang2024FitDiTAT}. Flow-based methods emphasize geometric alignment by estimating dense correspondences or flow fields to warp garment features onto target body images. GAN approaches employ adversarial learning to synthesize photorealistic try-on results, focusing on suppressing artifacts and enhancing garment--body fusion. Textual inversion methods introduce text-driven embeddings to adapt pre-trained diffusion models toward personalized garment--body synthesis. PBE methods adopt a CLIP encoder to guide a U-Net for transferring garment details from a reference image to the target body. Dual U-Net designs extend this approach by cascading two U-Nets, improving fine-grained texture recovery and alignment consistency. DiT methods utilize transformer-based diffusion architectures to achieve high-fidelity, scalable virtual try-on generation. Collectively, these paradigms trace the evolution of VTON models from early warping pipelines to modern diffusion--transformer frameworks, forming a comprehensive basis for studying cross-method generalization and robustness. 

\subsection{Source Data and Subset Construction}
The construction of these subsets primarily relies on the VITON-HD dataset~\cite{choi2021viton}, which contains 13,679 frontal-view images of women and top garments; following established protocols~\cite{lee2022high,choi2021viton}, we partition the dataset into 11,647 training and 2,032 testing images under an unpaired setting, reflecting practical scenarios where person and garment images are sourced independently, and standardize all images to $512\times384$ resolution for generation and evaluation. To further enrich diversity and evaluate robustness under more challenging conditions, we additionally incorporate the DressCode-Upper dataset~\cite{morelli2022dress}, which contains 15,363 upper-garment images; we select 1,800 images to form an additional unpaired test set, introducing greater discrepancies between garments and body images and thus providing a more demanding benchmark for cross-subset evaluation. Beyond public datasets, we augment garment diversity by collecting over 10,000 high-quality product images from e-commerce platforms, which broaden style and texture coverage for generating all six subsets. 


\subsection{Harmonization with Variant Backgrounds}

Most images in the VITON-HD and DressCode-Upper datasets feature plain or nearly white backgrounds, which substantially simplifies the virtual try-on detection task and fails to capture the complexity of real-world deployment scenarios. In practical applications, virtual try-on systems are expected to operate on user-generated content from social media or e-commerce platforms, where backgrounds are cluttered, contain rich textures, and often include complex lighting effects such as cast shadows or mixed color temperatures. To better reflect the practical difficulty and real-world relevance of virtual try-on detection, we enhance our benchmark by augmenting both generated try-on results and corresponding ground-truth images with complex, photorealistic backgrounds.

To achieve realistic augmentation, we adopt the \textit{IC-Light} framework~\cite{zhang2025scaling}, a diffusion-based illumination harmonization technique that enforces physically consistent light transport. Within our pipeline, each try-on image is directly processed by IC-Light, which automatically extracts the foreground and harmonizes it with a randomly sampled background from a diverse external dataset. This automated design eliminates the need for manual segmentation or parameter tuning, enabling large-scale generation of realistic and diverse try-on composites crucial for training and evaluating virtual try-on detection models.


For background sources, we employ the DIV2K dataset~\cite{agustsson2017ntire}, which contains 1000 high-resolution natural images covering diverse outdoor and indoor scenes such as urban landscapes, vegetation, architecture, and textured surfaces. By combining these backgrounds with IC-Light’s illumination-aware harmonization, we generate a large number of augmented images that preserve garment and body details while introducing realistic environmental complexity. The resulting benchmark, which we denote as VTONGuard, substantially enriches the diversity of virtual try-on detection tasks by simulating real-world deployment conditions. 

\subsection{Detection via Multi-Task Training}


For the VTON detection task, the image editing is mainly concentrated on the garment regions, making it particularly meaningful to introduce segmentation supervision over human-clothing areas. By explicitly modeling these regions, the network is encouraged to learn boundary-aware representations that capture subtle garment inconsistencies more effectively. Therefore, beyond the single-task classification setting, we extend ViT-based backbones into a multi-task (MT) framework by adding a segmentation head that predicts human-clothing masks in parallel with the real/fake classification branch. The joint training objective is defined as
\begin{equation}
\mathcal{L} = \mathcal{L}_{\text{cls}} + \mathcal{L}_{\text{seg}},
\end{equation}
where $\mathcal{L}_{\text{cls}}$ denotes the cross-entropy loss supervising the real/fake prediction, and $\mathcal{L}_{\text{seg}}$ is the binary cross-entropy loss supervising the predicted human-clothing masks. This multi-task design enforces boundary-aware representation learning and consistently yields improvements over the single-task variant across ViT-based methods. 

The segmentation masks are used only during training and are completely discarded at inference time, incurring no additional test-time cost. The supervision masks are pseudo labels generated by an off-the-shelf ATR human parsing model. Since all VTON methods adopt the same mask source, this identical supervision protocol guarantees a fair comparison without introducing bias from mask quality.

\begin{figure}[t]
    \centering
    \includegraphics[width=\linewidth]{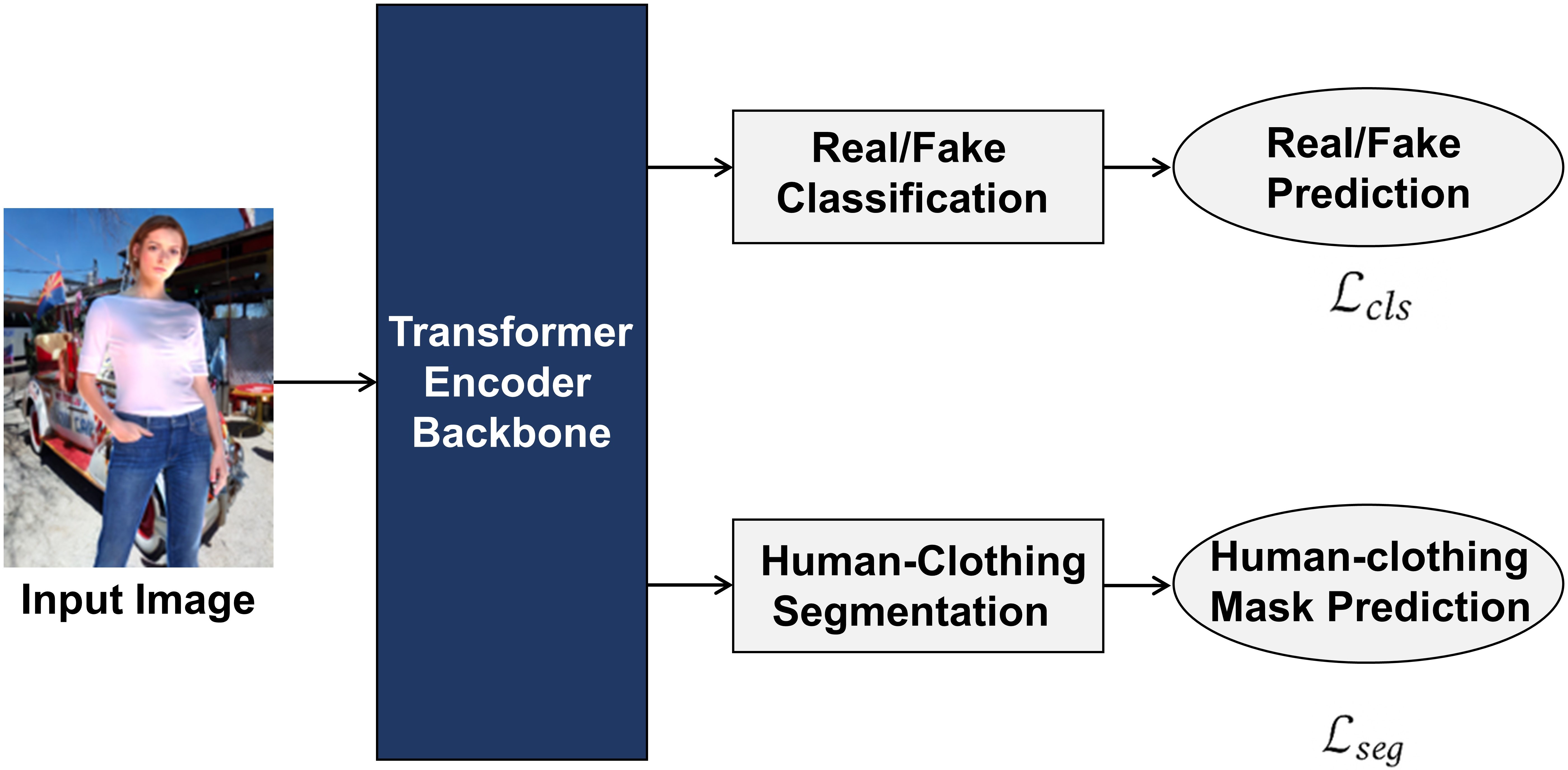}
    \caption{Overall architecture of our MT design. The model adopts a transformer encoder for feature extraction and adds a segmentation branch alongside the classification head to jointly predict real/fake labels and human-clothing region masks, enabling boundary-aware feature learning.}
    \label{fig:mitb2mt_arch}
\end{figure}

\begin{table*}[ht]
\centering
\caption{Per-subset detection performance (Acc/AP/AUC) of various detectors jointly trained on all subsets of VTONGuard.}
\label{tab:intra_dataset_comparison}
\resizebox{\linewidth}{!}{
\begin{tabular}{lcccccc|c}
\hline
\textbf{Model} & Flow-based & GAN & Textual Inversion & PBE & Dual U-Net & DiT & Average \\
\hline
ResNet-50 & 89.46/99.42/99.16 &76.91/85.53/85.89&66.08/65.13/71.69&64.82/63.73/69.69&73.99/80.57/82.25& 70.65/76.62/78.25 & 73.32/78.83/81.82       \\
Spec & 99.91/99.99/99.99 & 93.84/99.43/99.42 & 89.72/96.86/97.19 & 87.00/95.00/95.64 & 87.88/95.03/95.41 & 89.26/96.84/96.73 & 91.93/97.86/97.73 \\
\hline

Swin-T & 99.96/100.00/100.00 & 94.33/99.27/99.29 & 92.37/97.89/98.17 & 83.94/93.57/94.34 & 87.75/94.41/95.06 & 91.13/97.72/97.67 & 91.91/97.48/97.76 \\
Swin-T-MT &99.98/100.00/100.00 & 97.68/99.89/99.87 & 94.99/98.96/99.12 & 89.65/96.97/97.44 & 91.20/96.25/97.15 &93.61/98.73/98.74 & 94.52/98.47/98.72 \\
\hline
MiT-B2 & 99.92/100.00/100.00 & 96.71/99.97/99.97 & 97.81/99.81/99.84 & 97.65/99.78/99.81 & 94.13/98.56/98.66 & 96.29/99.49/99.49 & 97.08/99.60/99.63 \\
MiT-B2-MT
&100.00/100.00/100.00&98.65/99.96/99.96&97.84/99.79/99.82&97.08/99.59/99.67&95.14/99.19/99.18&96.64/99.59/99.59&97.89/99.68/99.70\\
\hline
\end{tabular}
}
\end{table*}

\begin{table*}[!t]
\centering
\caption{Cross-subset detection performance (Acc/AP/AUC) of MiT-B2-MT on VTONGuard. Each row corresponds to the training subset, and each column corresponds to the testing subset.}
\label{tab:mitb2_cross_dataset}
\resizebox{\linewidth}{!}{
\begin{tabular}{lcccccc|c}
\hline
\textbf{Train $\backslash$ Test} & Flow-based & GAN & Textual Inversion & PBE & Dual U-Net & DiT& Average \\
\hline
Flow-based &100.00/100.00/100.00&51.23/55.21/55.45&45.16/48.96/54.90&45.14/46.37/52.27&50.00/50.34/50.06&49.99/50.73/50.21&56.92/58.60/60.48
\\
GAN &50.00/66.26/68.90&99.84/100.00/100.00&45.37/53.79/59.19&44.98/52.60/57.11&50.22/61.61/60.75&50.39/62.79/62.98&56.80/66.18/68.16 \\
Textual Inversion   &50.43/79.25/75.33 &53.61/70.19/68.98 &97.75/99.76/99.79 &49.34/69.79/70.11 &59.84/80.67/78.92 &53.31/68.86/67.18 &60.88/78.75/76.72 \\
PBE      &48.56/55.86/52.89&54.22/64.52/63.62&47.21/58.16/63.40&97.54/99.68/99.74&54.37/75.07/74.84&51.20/63.44/63.29&58.85/69.46/69.63 \\
Dual U-Net     &50.21/79.54/78.74&61.08/83.99/84.17&54.49/56.47/67.55&51.77/70.92/77.08&91.07/98.55/98.64&78.74/95.43/96.69&64.56/80.48/83.15 \\
DiT        & 49.95/73.83/77.00  & 64.38/89.41/89.78 & 48.61/60.69/70.02 & 45.96/69.47/75.22 &68.72/95.28/94.82 & 96.32/99.73/99.73 & 62.32/81.40/84.43 \\
\hline
\end{tabular}
}
\end{table*}
\begin{table*}[!t]
\centering
\caption{Leave-one-subset-out detection performance (Acc/AP/AUC) of MiT-B2-MT on VTONGuard.}
\label{tab:remaining}
\resizebox{\linewidth}{!}{
\begin{tabular}{lcccccc|c}
\hline
\textbf{Model} & Flow-based & GAN & Textual Inversion & PBE & Dual U-Net & DiT & Average\\
\hline
MiT-B2-MT
&52.88/87.81/86.75&76.85/93.48/92.87&58.50/76.85/81.69&69.70/86.60/88.02&86.88/96.38/96.20&85.68/95.08/95.02&71.08/89.70/90.76\\

\hline
\end{tabular}
}
\end{table*}
\begin{table*}[!t]
\centering
\caption{Per-subset detection performance (Acc/AP/AUC) of MiT-B2-MT jointly trained on different reduced subsets of VTONGuard.}
\label{tab:tab4}
\resizebox{\linewidth}{!}{
\begin{tabular}{lcccccc|c}
\hline
\textbf{Model} & Flow-based & GAN & Textual Inversion & PBE & Dual U-Net & DiT & Average \\
\hline
Without Reduction
&100.00/100.00/100.00&98.65/99.96/99.96&97.84/99.79/99.82&97.08/99.59/99.67&95.14/99.19/99.18&96.64/99.59/99.59&97.89/99.68/99.70\\

Reduce to 1/2 & 99.94/100.00/100.00 &98.11/99.95/99.95&95.83/99.37/99.46&94.40/98.99/99.15& 93.99/98.76/98.77&94.84/99.03/99.04 & 96.18/99.35/99.39       \\

Reduce to 1/4 & 99.88/100.00/100.00 & 93.94/99.90/99.89 & 96.22/99.57/99.63 & 95.96/99.36/99.48 & 92.84/98.84/98.88 & 93.56/99.37/99.38 & 95.40/99.50/99.54 \\

Reduce to 1/8 & 99.96/100.00/100.00 & 96.12/99.92/99.92 & 94.41/98.79/98.91 & 92.23/97.78/98.05 & 93.46/98.54/98.55 & 93.33/98.63/98.63 & 94.91/98.94/99.00 \\

Reduce to 1/16 & 100.00/100.00/100.00 & 96.26/99.53/99.49 & 89.82/97.77/98.07 & 87.06/96.42/97.03 & 87.79/96.94/96.94 & 89.43/97.52/97.53 & 89.43/98.02/98.17 \\


\hline
\end{tabular}
}
\end{table*}

\section{Experiments}
\subsection{Implementation Details}
We comprehensively evaluate our benchmark using four representative detectors spanning convolutional and transformer-based architectures, covering spatial, frequency, and hierarchical modeling paradigms. ResNet-50~\cite{he2016deep} serves as a classical convolutional baseline due to its strong spatial feature extraction capabilities and efficiency. It is widely used in synthetic image detection and is fine-tuned here as a binary classifier to distinguish real from AI-generated try-on images, providing a reliable spatial-domain reference. Complementing this, Spec~\cite{zhang2019detecting} operates in the frequency domain by analyzing spectral representations to detect generative artifacts. By transforming images into the Fourier domain, Spec captures high-frequency inconsistencies introduced by upsampling, which are often invisible in the spatial domain, making it particularly robust against post-processing operations.

On the transformer side, we consider representative vision transformer backbones.
We adopt Swin Transformer-Tiny (Swin-T)~\cite{liu2021Swin}, a hierarchical vision transformer utilizing shifted window self-attention. This design significantly reduces computational cost compared to vanilla ViT while maintaining strong global and local modeling capabilities critical for detecting both coarse alignment errors and fine garment texture inconsistencies in VTON-generated images. We further include MiT-B2~\cite{xie2021segformer}, a hierarchical transformer architecture originally developed for semantic segmentation. MiT-B2 combines multi-scale feature extraction through a lightweight Mix Transformer encoder with efficient all-MLP decoders, capturing both global context and fine-grained structural details with low computational overhead. 

For evaluation metrics, we follow prior works on AI-generated image detection~\cite{wang2020cnn, wang2019detecting} and report accuracy (ACC) and average precision (AP) as primary indicators, with the classification threshold fixed at 0.5. We additionally report the Area Under the ROC Curve (AUC) to assess ranking quality across thresholds, providing a more comprehensive evaluation of detector performance in real-world settings where decision thresholds may vary. All experiments are conducted on a workstation equipped with eight NVIDIA RTX 4090 GPUs, ensuring consistent training conditions and enabling large-scale benchmarking across all subsets.

\subsection{Per-subset Analysis after Joint Training}

We train each detector on the full training set containing all subsets and evaluate it separately on each test subset to analyze detection difficulty across different try-on paradigms and the robustness of each model. By examining per-subset performance, we can identify which subsets pose greater challenges and which detectors generalize better across heterogeneous try-on methods. The per-subset results in Table~\ref{tab:intra_dataset_comparison} show that newer VTON methods produce more realistic images and are therefore harder to detect than earlier approaches. 

Spec consistently outperforms the ResNet-50 baseline, demonstrating the effectiveness of spectral features. Transformer-based detectors show even stronger performance, with the MiT-B2 family performing particularly well due to its ability to process high-resolution inputs and fuse multi-scale features. Moreover, our experiments demonstrate that introducing a multi-task (``MT'') design consistently benefits all ViT-based backbones. By adding a segmentation head that predicts human-clothing masks in parallel with real/fake classification, the network is explicitly guided to attend to garment-body boundaries and stitching regions, leading to boundary-aware representations that capture subtle editing artifacts. Across Swin-T and MiT variants, the MT extensions deliver consistent gains over their single-task counterparts, while introducing only negligible computational overhead. Among them, MiT-B2-MT achieves the strongest overall performance across all subsets, improving not only accuracy but also average precision and AUC. These results suggest that enhancing ViT-based detectors with segmentation-guided multi-task learning is broadly effective for VTON detection, and that future designs should focus on high-resolution input modeling and auxiliary supervision to remain robust as generative methods continue to advance.

\subsection{Cross-subset Generalization Analysis}

Cross-dataset generalization remains a persistent challenge in AIGC detection, where detectors trained on one generative paradigm often fail to transfer to others due to domain and artifact shifts. To examine whether this challenge also arises in our benchmark, we evaluate the best-performing detector, MiT-B2-MT, under a cross-subset setting: the model is trained on one subset (corresponding to a specific generation paradigm) and tested on all others. This protocol exposes the degree of feature overlap among paradigms and measures whether detectors can generalize beyond the paradigms seen during training.

As shown in Table~\ref{tab:mitb2_cross_dataset}, cross-subset performance drops considerably when training and testing paradigms mismatch, confirming that generalization gaps widely observed in generic AIGC detection benchmarks also persist in the virtual try-on setting. In contrast, training and testing on the same subset yields substantially higher performance, indicating that each subset contains paradigm-specific artifacts that detectors can easily exploit but fail to generalize across paradigms. Interestingly, subsets generated by newer paradigms exhibit stronger cross-paradigm transferability than earlier subsets, likely due to their higher visual fidelity and reduced reliance on low-level warping artifacts.

To evaluate generalization, we adopt a leave-one-subset-out protocol. For each target subset, we sample 1/5 of the training data from each of the other five subsets to form a mixed training set of comparable size. The model is trained on this mixed set and evaluated solely on the held-out subset, which represents a completely unseen generation paradigm. As shown in Table~\ref{tab:remaining}, training with diverse data from multiple VTON paradigms consistently improves generalization compared to single-subset training, despite using the same total data volume. This indicates that cross-paradigm diversity, rather than data scale, is the main driver of improved performance. The only exception is the flow-based subset, likely due to its lower resolution ($256\times192$), which differs substantially from the others.

\subsection{Data Reduction Experiments}
In this experiment, we progressively reduced the original training set to 50\%, 25\%, 12.5\%, and 6.25\% of its original size and evaluated the corresponding performance. Specifically, we randomly sampled a fixed proportion of data from each training set and merged them into a new training set for model training. As shown in Table~\ref{tab:tab4}, reducing the amount of training data led to a general decline in performance across most subsets, with the exception of Flow-based and GAN. For Flow-based, the high quality of its training samples may explain why even with less data, the performance remains strong. In the case of GAN, recognition performance appears less sensitive to training data reduction, indicating a degree of robustness to data sparsity. Despite the reduced data, the model still effectively distinguishes between VTON-generated and real images, which further demonstrates the robustness and superiority of our method. Additionally, some fluctuations in accuracy can be attributed to the heterogeneous quality within the dataset: random sampling may inadvertently select subsets with more high- or low-quality samples, influencing the results.

\section{Conclusion}
In this work, we introduced \textit{VTONGuard}, the first large-scale benchmark for detecting AI-generated virtual try-on (VTON) content in real-world settings. Comprising over 775,000 real and synthetic images with diverse poses, garments, and backgrounds, it captures the unique challenges of VTON detection where fine-grained garment-body alignment often renders synthetic images indistinguishable from real ones. Experiments show that transformer-based detectors outperform CNN counterparts, with our multi-task MiT-B2-MT achieving the best results by jointly optimizing classification and segmentation. For VTON detection model design, the introduction of multi-task learning proves particularly meaningful, as segmentation guidance helps detectors explicitly attend to garment-body boundaries. Cross-paradigm evaluations further reveal significant generalization gaps, underscoring the need for paradigm-agnostic representations for practical deployment. 
We hope this benchmark will advance robust VTON detection and practical applications.


\section{Acknowledgments}
This work was supported in part by the National Natural Science Foundation of China (Grant Nos. 62302295, 62595733, and 62561160155), the Shanghai Municipal Science and Technology Major Project (Grant No. 2021SHZDZX0102). This work was also supported by Ant Group.

\bibliographystyle{IEEEbib}
\bibliography{strings,refs}

\end{document}